\documentclass{article}
\usepackage[preprint]{neurips_2026}

\usepackage[utf8]{inputenc} 
\usepackage[T1]{fontenc}    
\usepackage{hyperref}       
\usepackage{url}            
\usepackage{booktabs}       
\usepackage{amsfonts}       
\usepackage{nicefrac}       
\usepackage{microtype}      
\usepackage{xcolor}         

\usepackage{gensymb}        
\usepackage{graphicx}       
\usepackage{longtable}
\usepackage{array}
\usepackage{listings}
\usepackage{amsmath}
\usepackage{comment}

\definecolor{codegreen}{rgb}{0,0.6,0}
\definecolor{codegray}{rgb}{0.5,0.5,0.5}
\definecolor{codepurple}{rgb}{0.58,0,0.82}
\definecolor{backcolour}{rgb}{0.95,0.95,0.92}

\lstdefinestyle{pythonstyle}{
    backgroundcolor=\color{backcolour},   
    commentstyle=\color{codegreen},
    keywordstyle=\color{blue},
    numberstyle=\tiny\color{codegray},
    stringstyle=\color{codepurple},
    basicstyle=\ttfamily\footnotesize,
    breakatwhitespace=false,         
    breaklines=true,                 
    captionpos=b,                    
    keepspaces=true,                 
    numbers=left,                    
    numbersep=5pt,                  
    showspaces=false,                
    showstringspaces=false,
    showtabs=false,                  
    tabsize=2,
    language=Python
}
\title{Insulin4RL:\@ Real-Time Insulin Management in the Intensive Care Unit for Offline Reinforcement Learning}

\author{%
  Thomas Frost\thanks{Corresponding author} \\
  Institute of Health Informatics\\
  University College London\\
  London, United Kingdom\\
  \texttt{thomas.frost.21@alumni.ucl.ac.uk}\\
  \And
  Steve Harris\\
  Institute of Health Informatics\\
  University College London\\
  London, United Kingdom\\
  \texttt{steve.harris@ucl.ac.uk}\\
}

\begin{document}

\maketitle

\begin{abstract}

  Offline reinforcement learning (ORL) offers the potential to improve the quality of clinical decision-making using historical electronic health record (EHR) data. Current training and evaluative practices in this field rely heavily on EHR datasets that have been temporally discretised into fixed, regular time intervals. Discretisation creates fictional representations of complex clinical scenarios and compromises the generalisability of retrospective model evaluations. In this paper, we introduce \textbf{Insulin4RL}, a healthcare ORL dataset featuring naturally irregular inputs and actions from real clinical trajectories. Derived from MIMIC-IV, Insulin4RL comprises over 375,000 labelled decisions across 12,209 patients requiring insulin infusion titration in the Intensive Care Unit. The dataset can thus be used for research into ORL model performance under realistic clinical sampling assumptions. We provide a description of the dataset's structure and characteristics, baseline performance metrics using model-free offline reinforcement learning, and a standardised evaluation protocol using fitted Q-evaluation. We conclude with suggested areas for future research that could be addressed using this resource.
\end{abstract}

\section{Introduction}

Reinforcement learning holds significant potential for optimising and personalising clinical decision-making~\citep{liu2020reinforcement, jayaraman2024primer}. To minimise patient risk, these algorithms are typically trained and evaluated using retrospective electronic health record (EHR) data---known as batch or offline reinforcement learning (ORL)~\citep{levine2020offline}. With a scarcity of prospective clinical trials due to ethical and practical barriers~\citep{wang2023optimized, gao2023offline, fan2026reinforcement}, off-policy evaluation (OPE) has become the de facto standard for assessing model safety and efficacy~\citep{gottesman2018evaluating, tang2021model}. Consequently, the scientific conclusions drawn about healthcare ORL models depend overwhelmingly on the design and implementation of these retrospective evaluation pipelines.

However, OPE may systematically mask suboptimal model behaviours. In particular, when datasets are no longer reflective of the intended environment, the model could learn maladaptive policies that seem beneficial during retrospective analysis but are harmful at deployment. A prevailing research practice involves temporally discretising EHR datasets by aggregating irregularly sampled clinical events into fixed-length windows (e.g., 4 hours)~\citep{lipton2016modeling, sun2025exploring, frost2026hidden}. While regular clinical decision labels may fit naturally within the standard Markov decision process (MDP) framework~\citep{bellman1957markovian, sutton2018reinforcement}, evaluations conducted on discretised data create a smoothed, fictional representation of complex clinical realities. In doing so, they are prone to significant bias~\citep{schulam2018discretizing, jeter2019artificial, tallec2019deep, frost2026hidden} and uncertainty over the `right' window size to use~\citep{lu2020deep, wu2023value, sun2025exploring}. If the evaluation environment deviates sufficiently from the naturally irregular reality of clinical care, the resulting claims about model performance and safety may in turn be fundamentally compromised.

To advance the progress of ORL training and evaluation in healthcare, we introduce \textbf{Insulin4RL}: a readily curated EHR dataset explicitly designed to study and evaluate model performance under realistic clinical sampling assumptions. Derived from the MIMIC-IV dataset~\citep{johnson2023mimic}, Insulin4RL focuses on the titration of intravenous insulin infusions for critically unwell patients in the Intensive Care Unit (ICU)---a continuous-time control problem for optimising irregularly sampled blood glucose measurements~\citep{plummer2014dysglycaemia, desgrouas2023insulin, adie2023association}, with a lack of clarity over the ideal approach for different patient subgroups~\citep{wilson2007intensive, nice2009intensive, bohe2021individualised, plummer2022time, gunst2023tight}. By preserving the natural frequency of sampled inputs and decisions, Insulin4RL moves the emphasis towards realistic timings under a semi-Markov decision process (SMDP) for evaluating these models.

\begin{figure}[t]\label{fig:consort}
  \centering
  \includegraphics[width=0.6\textwidth]{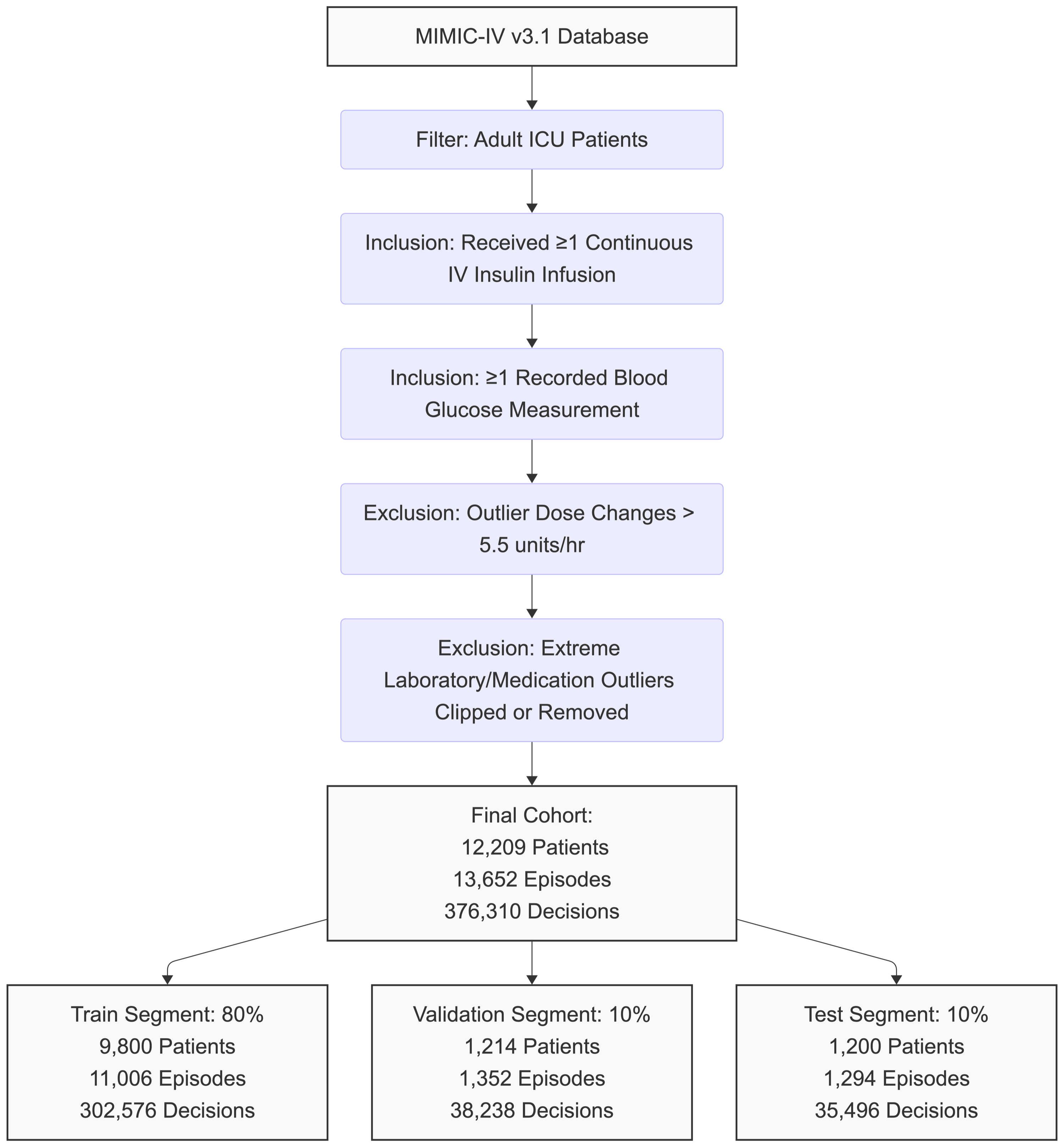}
  \caption{\textbf{CONSORT diagram:} Patients in Insulin4RL are derived from a cohort of Intensive Care Unit admissions in the MIMIC-IV v3.1 dataset. Each episode must have at least one continuous intravenous insulin infusion with at least one associated blood glucose measurement. Outlier inputs are clipped or excluded according to percentile values. Decision labels are also manually restricted to a clinically appropriate adjustment range ($\pm 0-5.5$ units/hr).}
\end{figure}

We provide the dataset in an immediate-use format for reinforcement learning along with template experiments for training and evaluating ORL models in this setting. Through baseline experiments with behavioural cloning, implicit Q-learning, and conservative Q-learning, we demonstrate how differing temporal assumptions can lead to divergent policies in this dataset. Insulin4RL can thus be used not only as a training resource, but as an evaluative framework for assessing the robustness of model behaviours to irregularly timed and unpredictable decision-making, in order to provide more reliable claims about ORL safety in this field.

\section{Related work}

One of the most influential evaluation benchmarks for healthcare-based ORL has been the MIMIC-III sepsis cohort derived by \citet{raghu2017continuous}. This dataset aggregates a range of input features and actions (vasopressor and intravenous fluid doses) into discrete 4-hour windows and has inspired numerous subsequent works~\citep{komorowski2018artificial, tang2020clinician, roggeveen2021transatlantic, fatemi2021medical, wu2023value, tu2025offline}. Later research expanded on this dataset: \citet{kuo2022health} released synthetic benchmark datasets for hypotension (1-hour windows), sepsis (4-hour windows), and HIV (1-month windows), while the recent MIMIC-Sepsis benchmark updated the sepsis cohort using MIMIC-IV data with the same 4-hour aggregate bins~\citep{johnson2023mimic, huang2025mimic}. Whilst these datasets have catalysed a large volume of ORL research, the use of discretisation has inadvertently become a \textit{de facto} practice that limits the interpretability of their results in a continuous-time clinical context.

In the domain of glycaemic control, the OhioT1DM dataset provides eight weeks of continuous glucose monitoring, insulin pump rates, and meal data for outpatients with type 1 diabetes~\citep{marling2020ohiot1dm}. Although primarily intended for glucose prediction, it frequently serves as an evaluation benchmark for RL-driven continuous ``artificial pancreas'' systems~\citep{zhu2023offline}. However, this data reflects high-frequency actions in stable community patients, contrasting with the sporadic, clinician-queried interventions of the ICU. Recent literature has increasingly explored offline RL for inpatient insulin titration to manage critical illness~\citep{wang2023optimized, desman2025distributional}, although public benchmark resources remain scarce. \citet{robles2021data} released a glucose management dataset derived from MIMIC-III for ICU patients; however, the dataset is intended for descriptive use (limiting features to only insulin and glucose) and lacks the comprehensive state representation found in our dataset.

\begin{figure}[t]\label{fig:umap}
  \centering
  \includegraphics[width=1.0\textwidth]{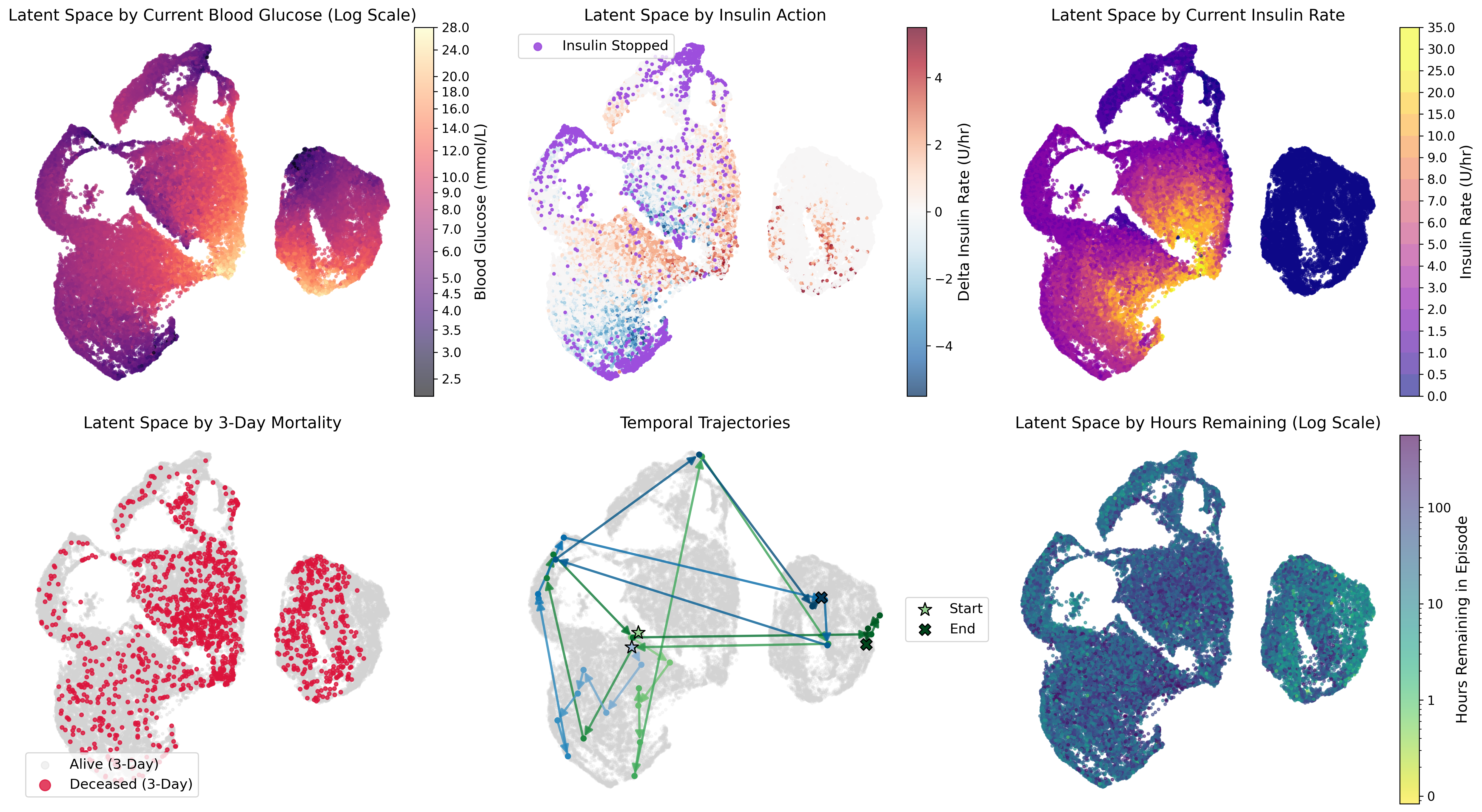}
  \caption{\textbf{Visualisation of the learned state state space using UMAP projection.} The top row and bottom-right panel show clear gradients for blood glucose, insulin rates/actions taken, and remaining duration of the episode. The bottom-left panel separates out surviving and non-surviving patients. The bottom-center panel maps the evolution of two sample episodes from initial (star) to final (cross) state.}
\end{figure}

\section{Methods}

\subsection{Preliminaries}

The standard reinforcement learning framework is the Markov decision process (MDP)~\citep{bellman1957markovian, sutton2018reinforcement}, defined by the tuple $\{\mathcal{S}, \mathcal{A}, \mathcal{P}, \mu_0, R, \gamma\}$ containing the state space $s_t \in \mathcal{S}$; the action space $a_t \in \mathcal{A}$; the transition probability $\mathcal{P}$ for reaching state $s_{t+1}$ given state $s_t$ and action $a_t$; the initial state distribution $\mu_0$; the reward function $\mathcal{R}$ following a transition $(s_t,a_t,s_{t+1})$; and the discount factor $\gamma$ for future rewards. Assuming the future only depends on the current state and action (the Markov property), the goal is to find a policy $\pi(a|s)$ that maximises the expected discounted return $G_t = \sum_{k=0}^\infty \gamma^k r_{t+k}$. This framework inherently assumes that transitions occur over fixed, regular time steps and that the true state is fully observable.

However, in healthcare settings, the underlying physiological state is latent and interventions can occur at unpredictable intervals. To provide a more rigorous framework, we can instead model the environment as a finite-horizon partially observable semi-Markov decision process (SMDP)~\citep{white1976procedures, yu2006approximate, zhang2016continuous}. We extend the standard tuple to include $\{\mathcal{O}, \Omega, \eta\}$, where $\textbf{o}_t \in \mathcal{O}$ represents the set of noisy, irregularly sampled observations in the EHR; $\Omega(\cdot|s,a)$ is the observation probability function; and $\eta \in \mathbb{N^+}$ is the variable sojourn time (in minutes) before the next independent decision. Consequently, the transition function $\mathcal{P}(\cdot|s,a)$ yields a joint distribution over the next state and the sojourn time $\eta$, and the discount factor $\gamma$ is dynamically adjusted to $\gamma^\eta$ for each interval. Because the analytical solution for $\Omega$ is intractable for continuous EHR data, we recommend the use of a sequential model (e.g., recurrent neural network) to infer a latent representation of $s_t$ from the observation-action history $h_t = (\textbf{o}_0, \textbf{o}_1, \ldots, a_0, a_1 \ldots)$ \citep{hausknecht2015deep, igl2018deep, ni2022recurrent}.

\begin{table}
  \caption{Baseline demographics, admission types, and highest-prevalence comorbidities of the Insulin4RL patient cohort. Continuous variables are presented as means, and categorical variables as percentages.}\label{tab:demographics}
  \centering
  \begin{tabular}{lr  lr}
    \toprule
    \textbf{Characteristic} & \textbf{Value} & \textbf{Comorbidity} & \textbf{Prevalence (\%)} \\
    \midrule
    \multicolumn{2}{c}{\textbf{Demographics \& Admission}} & Hypertension (Uncomplicated) & 50.9 \\
    Age (years), Mean & 64.4 & Cardiac Arrhythmias & 48.9 \\
    Weight (kg), Mean & 85.2 & Fluid and Electrolyte Disorders & 47.2 \\
    Female (\%) & 34.1 & Valvular Disease & 37.7 \\
    Male (\%) & 65.9 & Diabetes (Uncomplicated) & 32.8 \\
    Emergency Admission (\%) & 66.8 & Congestive Heart Failure & 29.3 \\
    Elective Admission (\%) & 33.2 & Diabetes (Complicated) & 25.5 \\
    \cmidrule{1-2} 
    \multicolumn{2}{c}{\textbf{Reported Ethnicity (\%)}} & Hypertension (Complicated) & 24.7 \\
    White & 68.3 & Coagulopathy & 24.7 \\
    Black & 9.2 & Renal Failure & 22.3 \\
    Hispanic & 4.0 & Chronic Pulmonary Disease & 21.8 \\
    Asian & 2.5 & Obesity & 17.2 \\
    Other / Unknown & 16.0 & Peripheral Vascular Disorders & 16.0 \\
    \bottomrule
  \end{tabular}
\end{table}

\subsection{Data source and patient cohort}

Insulin4RL is derived directly from de-identified healthcare data in the Medical Information Mart for Intensive Care (MIMIC-IV v3.1) dataset~\citep{johnson2023mimic}, available on the PhysioNet platform~\citep{johnson2024mimic, goldberger2000physiobank}. No additional Institutional Review Board (IRB) approval was required for this work.

Under the guidance of clinical domain experts, we identified a cohort of adult ICU patients who had received at least one continuous intravenous insulin infusion with at least one recorded blood glucose measurement. To capture the full context of clinical decision-making, an evaluation episode (or trajectory) is defined as the continuous period surrounding an infusion (including intermittent discontinuations), padded with 24 hours of pre- and post-infusion data. An infusion is considered terminated if, at cessation, the infusion is not restarted again for at least 48 hours. To promote robust training and evaluation practices, the cohort is pre-partitioned into an 80/10/10 split for training, validation, and testing at the patient level.

\subsection{State representation and input features}

Inspired by the MEDS format~\citep{arnrich2024medical}, the patient state is represented as a raw sequence of retrospective medical event tuples $(f, t, v)$, where $f$ is the feature code; $t$ is the elapsed time (in minutes) prior to the current decision; and $v$ is the numerical value. Individual events encompass 140 possible demographic, laboratory, medication, and physiological features (detailed in Appendix Table~\ref{tab:feature-summary}). The sequence length is capped at 400 events and events older than 7 days are excluded. The sequence is sorted from old to new (beginning with the patient's age, sex, and weight), and laboratory events are also sorted by novelty (so that rare features are not excluded in favour of high-frequency features). Lastly, any currently active drug infusions are repeated at the end of the sequence (with a relative time $t=0$), to ensure all ongoing infusions are never truncated from the sequence.

Where relevant, medications are separated into `bolus' (discrete one-off doses lasting one minute or less) and `rate' values (commencing of an infusion lasting longer than one minute at a fixed rate, or any subsequent changes). Overlapping infusions of the same drug are merged into a net infusion rate. Antibiotic doses are represented as binary `on/off' bolus events. Extreme outlier values were conservatively clipped (for medications) or excluded (for laboratory results), using the 0.1/99.9 and 99.5 percentile values respectively (derived exclusively from the training cohort).

\subsection{Action space and decision labels}

A primary difficulty in curating irregular EHR data for ORL is the absence of explicit maintain or ``do nothing'' action labels~\citep{zhang2016continuous}. To address this whilst grounding the labels in realistic clinical workflows, we labelled decisions based on when a clinician is \textbf{most likely to query a model} for dosing advice.

In critical care, insulin titration is explicitly driven by blood glucose measurements. Therefore, we define a decision point at every recorded blood glucose measurement within an episode~\citep{zhang2021identifying}. By analysing the period from 5 minutes before to 30 minutes after each check---or until 5 minutes before the next check, whichever is sooner---we extract one of three mutually exclusive, clinically meaningful categorical actions:

\begin{itemize}
  \item \textbf{Maintain infusion:} Insulin rate stays stable (<0.25 units/hr change).
  \item \textbf{Stop infusion:} Insulin rate drops from >0.1 units/hr to <0.1 units/hr.
  \item \textbf{Change infusion:} Insulin rate is adjusted by >0.25 units/hr (without stopping).
\end{itemize}

We also include the raw \textbf{Delta change} difference in units/hr between the new and old infusion rates. Changes by more than 5.5 units/hr are treated as outlier events and removed from the pool of labelled decisions, in order to limit models to safer adjustment ranges.

\subsection{Evaluative signals and rewards}

To support a variety of evaluation setups, each labelled transition includes multiple potential reward signals: the current and subsequent blood glucose measurements, survival indicators at multiple horizons (1, 3, 7, 14, and 28 days), and the variable sojourn time $\eta$ (in minutes) to the next state.

\subsection{Dataset structure and availability}

To ensure long-term accessibility, Insulin4RL is hosted on PhysioNet~\citep{goldberger2000physiobank} at \url{https://physionet.org/content/Insulin4RL}. The dataset is provided in two key formats: 

\begin{itemize}
  \item \texttt{all\_data.parquet}: An Apache Parquet DataFrame containing all raw input and labelled feature data.
  \item \texttt{*.safetensors}: A collection of tensor transitions (states, actions, rewards, etc.). Unlike the DataFrame, all states here have been log-transformed and standardised using values derived from the training cohort of patients.
\end{itemize}

The dataset includes several metadata files (e.g., feature mappings, outlier thresholds, demographics), as well as an accompanying Jupyter notebook for reproducing all experiments in this paper. All code to reproduce the dataset is available at \url{https://github.com/tdgfrost/insulin4rl}. 

\section{Experiments}

\begin{figure}[h]
  \centering
  \includegraphics[width=0.95\textwidth]{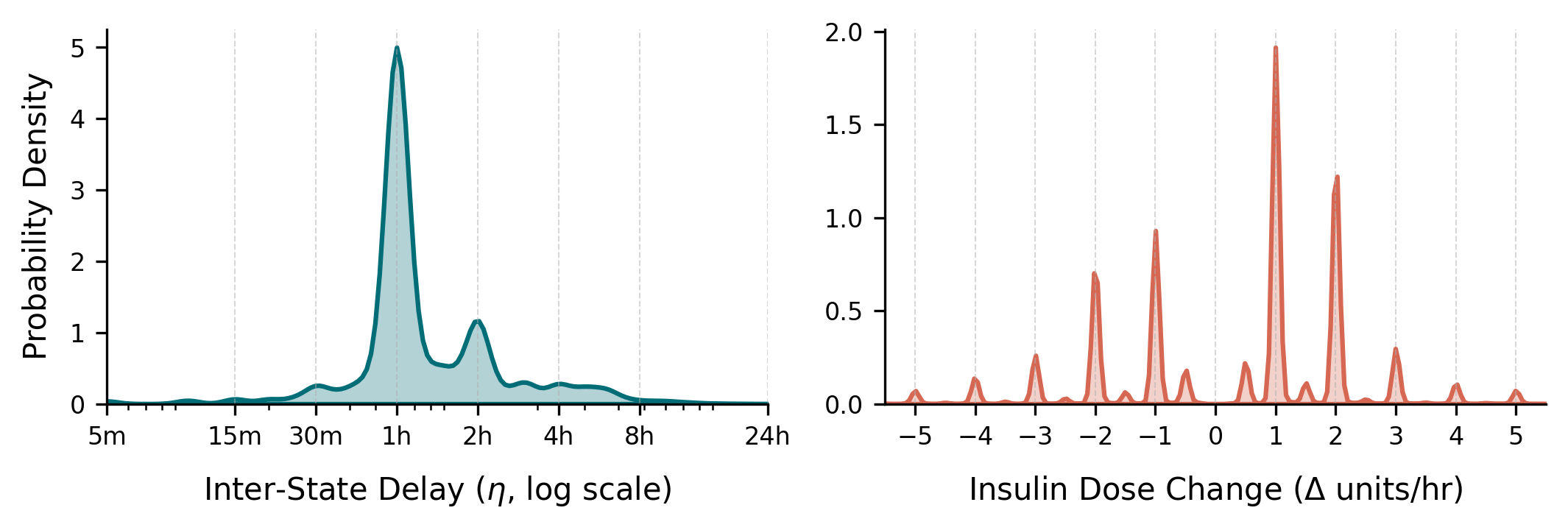}
  \caption{\textbf{Distribution of temporal and action characteristics in Insulin4RL.} (Left) The distribution of naturally irregular intervals ($\eta$) between adjacent states in the dataset. (Right) The distribution of insulin dose changes ($\Delta$ units/hr), demonstrating a clear clinical tendency towards discrete dose increments.}
  \label{fig:action-distribution}
\end{figure}

To demonstrate the utility of Insulin4RL as a training and evaluative benchmark, we provide a suite of baseline experiments. Rather than seeking state-of-the-art performance, these experiments are designed to validate the dataset's clinical fidelity and illustrate how differing temporal evaluation assumptions (MDP vs SMDP) may impact policy learning.

\subsection{Model architecture and action space}

All models process the irregularly sampled input sequences using a shared sequence-modelling architecture (a feature-embedding layer followed by 1D-CNN and LSTM layers, detailed in Appendix~\ref{app:model_code})~\citep{tipirneni2022self, kim2014convolutional,hochreiter1997long}.

A common problem in learning continuous healthcare interventions is action representation. Based on the clinical distribution of insulin adjustments (Figure~\ref{fig:action-distribution}), each policy model $\pi_\psi$ outputs two categorical heads. The first dictates the discrete action (``Do nothing'', ``Stop the infusion'', and ``Change the infusion rate''). The second predicts the magnitude of the change using 14 ordered dose classes~\citep{frank2001simple}, defined by the ascending set $\mathcal{V} = \{\pm 5, \pm 4, \pm 3, \pm 2, \pm 1.5, \pm 1, \pm 0.5\}$ units/hour. This formulation achieves an $R^2$ value of 99.9\% against the true continuous dose changes in the dataset, with a mean absolute error of 0.017 units.

Rather than standard classification, the network outputs 13 threshold probabilities $p_j = P(\Delta D \geq v_{j+1})$. The probability of selecting an exact dose class $k$ is computed as the joint probability of satisfying only the first $k-1$ thresholds:

\begin{equation}
  P(C = v_k) = \frac{1}{Z} \left( \prod_{j=1}^{k-1} p_j \right) \left( \prod_{j=k}^{13} (1 - p_j) \right)
\end{equation}

where $Z$ is a normalization constant. In contrast, critic models $Q(s,a)$ use a single output head encompassing all 16 possible actions.

\subsection{Validating clinical fidelity via behavioural cloning}

In order to verify that the dataset contains sufficient signal for learning, we first trained a behavioural cloning (BC) policy~\citep{kumar2022should} to mimic clinician behaviour using standard cross-entropy and ordinal regression losses. We then assessed the coherence of the learned state space by performing inference on the validation cohort and projecting the penultimate hidden state into a two-dimensional space using Uniform Manifold Approximation and Projection (UMAP), with n\_neighbors=10 and min\_dist=0.3 \citep{mcinnes2018umap}.

\begin{table}[h]
\centering
\caption{\textbf{Behavioural cloning performance metrics:} Results in the validation cohort of patients after training a behavioural cloning model. Categorical actions are reported using area under the receiver operating characteristic (AUROC) curve. Dosing is reported using mean absolute error (MAE) between the model recommended dose and actual observed dose (in units per hour).}
\label{tab:cloning-auroc-results}
\begin{tabular}{lcc}
\toprule
Action & Metric & Score \\
\midrule
Do nothing & AUROC & 0.824 \\
Stop the infusion & AUROC & 0.942 \\
Decrease the infusion rate & AUROC & 0.912 \\
Increase the infusion rate & AUROC & 0.897 \\
\midrule
Dose accuracy & MAE & 0.611 \\
\bottomrule
\end{tabular}
\end{table}

\subsection{Evaluating offline reinforcement learning using off-policy evaluation}

To establish baseline RL evaluations on the dataset, we trained policies using two state-of-the-art model-free ORL algorithms: implicit Q-learning (IQL)~\citep{kostrikov2022offline} and conservative Q-learning (CQL)~\citep{kumar2020conservative}. Two variants of each algorithm were trained: one using the standard MDP framework (with a fixed discount factor $\gamma=0.95$) and another using the SMDP framework (adjusting the discount factor dynamically to $\gamma^\eta$, using $\gamma=0.999$ per minute). For IQL, the recommended action is the argmax policy action (and dose if relevant). For CQL, the recommended action is the argmax action for the critic Q-function.

For all RL baselines, the reward $R_t$ evaluates the subsequent blood glucose level $G$ (in mg/dL) using a transformed Magni risk index~\citep{magni2007model}. We apply an asymmetric transformation $f(G) = 1.509\left[(\ln G)^{1.084} - 5.381\right]$ to heavily penalise hypo- or hyperglycaemia, defining the final reward as $R_t = \max (0.051 - 0.1 f(G)^2, 0.51 - f(G)^2)$.

Models are evaluated using a standardised off-policy evaluation methodology called Fitted Q-Evaluation (FQE)~\citep{le2019batch}. FQE was selected due to its robust performance in high-dimensional continuous state spaces where the underlying behavioural policy is unknown and complex~\citep{voloshin2021empirical, tang2021model}. We train dedicated FQE models to predict the expected return $\mathbb{E}_{a' \sim \pi_e}\left[Q(s_0, a')\right]$ for each of the learned evaluation policies $\pi_e$ (BC, IQL, CQL), as well as a baseline FQE model trained via SARSA to evaluate the empirical dataset behaviour.

Additional implementation details can be found in Appendix~\ref{app:implementation_details}.

\subsection{Computational requirements}
Insulin4RL was generated on an Macbook M1 Pro laptop and took 7 minutes for initial MIMIC-IV file conversions and 5 minutes for dataset generation. Jupyter notebook experiments were performed on a Linux desktop using an NVIDIA TITAN RTX GPU and took the following time to complete: 80 minutes (behavioural cloning), 3.5 hours (implicit Q-learning), 80 minutes (conservative Q-learning), and 4 hours (fitted Q-evaluation).

\section{Results}

\subsection{Cohort characteristics}

Insulin4RL contains complex clinical trajectories for 13,652 distinct insulin infusion episodes across 12,209 unique ICU patients, with 376,310 naturally timed decision points. The cohort (Table~\ref{tab:demographics}) features a high prevalence of critical comorbidities, with the majority (66.8\%) representing unplanned emergency admissions. Overall 28-day survival for the cohort is 91.4\%. As detailed in Appendix Table~\ref{tab:episode-stats}, episode durations show a significant rightward skew (median 28 hours, maximum 35 days). This provides models with a diverse mix of short-term and prolonged ICU stays.

\subsection{Validating the state space representation}

Figure~\ref{fig:umap} visualises the learned latent representations from the behavioural cloning model using UMAP. The projection displays clear, organised gradients corresponding to meaningful physiological patterns, including current blood glucose, prescribed insulin rates, and remaining duration of the episode. Notably, the latent space automatically disentangles survival outcomes (despite survival not being a cloning objective) and provides clear temporal pathways for tracking patient trajectories. This degree of coherence helps to confirm the extent to which Insulin4RL captures meaningful and complex clinical dynamics within this environment.

\subsection{Evaluating cloning and ORL model performance}

\begin{figure}[t]
  \centering
  \includegraphics[width=0.95\textwidth]{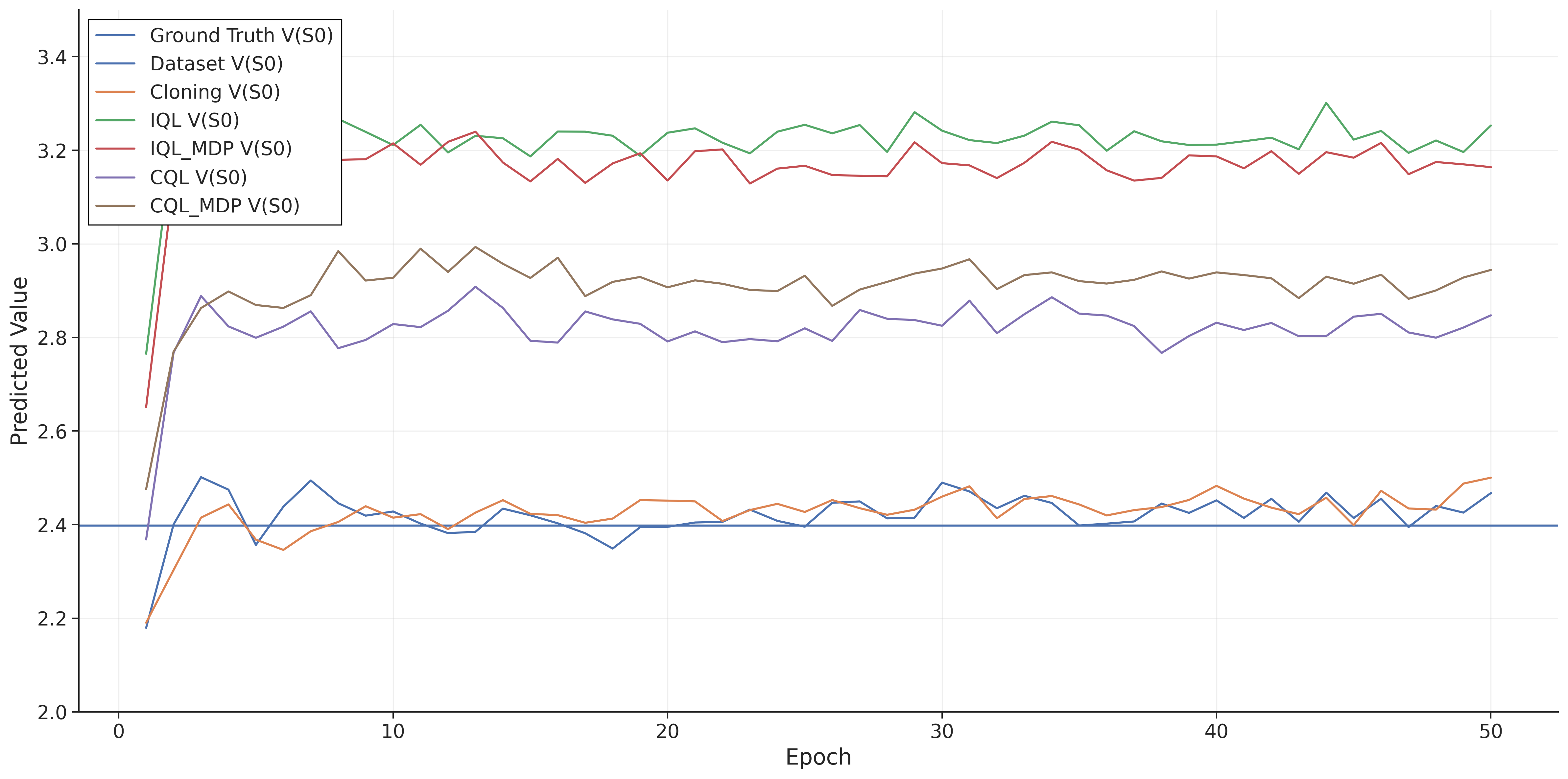}
  \caption{\textbf{Fitted Q-evaluation (FQE) results for cloning and offline reinforcement learning models:} FQE models are individually fitted to the dataset behaviour, behavioural cloning, and IQL/CQL models (both MDP and SMDP variants). The FQE models trained on the dataset and cloning policies are well-calibrated against the true discounted return. IQL and CQL policies improve on this baseline, with intra-algorithmic differences depending on whether the policy was trained under MDP or SMDP assumptions.}
  \label{fig:fqe-results}
\end{figure}

Table~\ref{tab:cloning-auroc-results} establishes the baseline ability of the behavioural cloning model to reproduce observed clinician behaviour, achieving high discrimination for active interventions (AUROC 0.89--0.94) and predicting continuous dose adjustments with reasonable precision (MAE of 0.61 units/hr). The slightly lower performance for the ``do nothing'' action suggests that inaction may be characterised by greater degrees of interclinician variability compared to the more active interventions.

Off-policy evaluation using FQE assesses the ability of established ORL algorithms to improve on this baseline behaviour. The FQE model calibrated against the empirical dataset behaviour shows a close correlation with the true expected return of the dataset, as does the FQE model trained against the behavioural cloning model. All ORL models had the capacity to exceed this baseline, with IQL algorithms trained and evaluated under an SMDP framework achieved the best overall returns. The diverging returns between MDP and SMDP for both IQL and CQL models highlights the sensitivity of these methods to the underlying temporal assumptions, emphasising the value of a naturally timed evaluation dataset.

\section{Discussion}

In this work, we introduced \textbf{Insulin4RL}, a large-scale training and evaluation resource designed to challenge the commonplace use of temporally discretised datasets in healthcare-based offline reinforcement learning. With over 375,000 naturally timed decision points for ICU insulin titration, we aim to shift the focus of ORL evaluation towards more realistically timed clinical data. Our baseline experiments successfully demonstrate the integrity of Insulin4RL as a collection of strong, coherent clinical signals. Even when trained against cloning objectives, UMAP visualisation shows natural clustering of data points around non-glucose-related information (e.g., mortality, duration of episode). 

For off-policy evaluation using FQE, we show that even identical model architectures (with identical inputs and labels) can learn differently optimal behaviours under standard MDP or semi-MDP frameworks. This underscores our concern that these models are highly sensitive to temporal assumptions, and that the widespread use of discretised datasets with evenly spaced decision intervals may contribute towards a general overestimation of model safety and efficacy in real-world applications. In particular, it supports our general view that the health ORL field would benefit from more evaluation benchmarks that accurately represent decision timing in clinical environments. Our principal goal is that the adoption of continuous-time evaluation frameworks will promote an era of more rigorous and realistic ORL model assessment---bringing the field one step closer to real-world trials.

\paragraph{Limitations} 
While Insulin4RL offers a more realistic evaluation environment, it inherits certain limitations from the MIMIC-IV parent dataset: derived from a single US institution, the data (and any evaluated policies) may not generalise well to other hospitals or countries with different practices or patient demographics, and could risk entrenching a set of conclusions that are not representative for many patients outside of this cohort. It is for this same reason that models trained on this dataset may not be easily validated on other datasets (and vice versa). Additionally, observational EHR data is naturally subject to unmeasured confounding (e.g., visual cues at the bedside), which could lead to causally confusing signals for the model. Finally, while FQE is a well-established OPE technique, it is susceptible to out-of-distribution extrapolation errors; we have not yet benchmarked alternative OPE methodologies, such as doubly robust estimation~\citep{jiang2016doubly}.

\paragraph{Future Directions}
Insulin4RL opens several avenues for future research. First, the dataset serves as an ideal testbed for developing novel continuous-time ORL algorithms. Second, researchers can utilize this dataset to benchmark new OPE methodologies in a high-dimensional, healthcare-specific setting with realistically unpredictable decision intervals. Finally, future works could explore tailored reward functions and potentially identify subgroup-specific policies for insulin titration, serving as a basis for future research in insulin management for ICU patients.

\section{Conclusion}

The scientific consensus surrounding healthcare ORL is ultimately limited by the quality of our evaluation practices. Currently, the field's reliance on temporally discretised benchmarks may not sufficiently identify how models will perform in stochastic, real-world clinical environments. Insulin4RL directly addresses this vulnerability by providing a collection of more than 375,000 naturally timed decisions for ICU insulin management. We show that even in the setting of identical inputs and labels, differences in temporal assumptions can affect the behaviours learned by ORL models. We release this dataset and accompanying tutorial notebook to encourage the community to develop more robust, clinically realistic ORL models that can safely handle the stochastic nature of real-world clinical decision-making.

\begin{ack}
TF was funded by the Engineering and Physical Sciences Research Council (EPSRC) as part of the UK Research and Innovation (UKRI) Centre for Doctoral Training in AI for Healthcare (grant EP/S021612/1) and supervised by SH at the National Institute for Health and Care Research (NIHR) University College London Hospitals (UCLH) Biomedical Research Centre (BRC)\@. Funders played no role in study design, data collection, analysis and interpretation of data, or the writing of this manuscript. The views expressed in the text are those of the authors and not necessarily those of the funders. There are no competing or conflicts of interest. We also gratefully acknowledge the facilities provided by University College London (UCL), which helped enable this research. We thank the team at PhysioNet for publishing and hosting the dataset, and the team behind MIMIC-IV, which enabled the creation of this resource. Finally, we thank Jack Parker-Holder for suggesting the publication of this dataset, and Simon Ellershaw for his valuable input into the written manuscript.
\end{ack}

\bibliographystyle{plainnat}
\bibliography{references}

@string{neurips = "Advances in Neural Information Processing Systems"}

@string{icml    = "International Conference on Machine Learning ({ICML})"}

@string{iclr    = "International Conference on Learning Representations ({ICLR})"}

@Article{komorowski2018artificial,
  author         = {Komorowski, Matthieu and Celi, Leo A and Badawi, Omar and Gordon, Anthony C and Faisal, A Aldo},
  title          = {The {A}rtificial {I}ntelligence {C}linician learns optimal treatment strategies for sepsis in intensive care},
  journal        = {Nature Medicine},
  year           = {2018},
  month          = nov,
  volume         = {24},
  number         = {11},
  pages          = {1716--1720},
  doi            = {10.1038/s41591-018-0213-5},
}

@Article{kumar2022should,
  author         = {Kumar, Aviral and Hong, Joey and Singh, Anikait and Levine, Sergey},
  title          = {When should we prefer offline reinforcement learning over behavioral cloning?},
  year           = {2022},
  journal        = {arXiv:2204.05618},
  doi            = {10.48550/arXiv.2204.05618},
}

@article{schulam2018discretizing,
  author         = {Schulam, Peter and Saria, Suchi},
  title          = {Discretizing Logged Interaction Data Biases Learning for Decision-Making},
  journal        = {arXiv preprint arXiv:1810.03025},
  year           = {2018},
  doi            = {10.48550/arXiv.1810.03025},
  url            = {https://arxiv.org/abs/1810.03025},
}

@article{jeter2019artificial,
  author         = {Jeter, Russell and Josef, Christopher and Shashikumar, Supreeth and Nemati, Shamim},
  title          = {Does the "{Artificial Intelligence Clinician}" learn optimal treatment strategies for sepsis in intensive care?},
  journal        = {arXiv preprint arXiv:1902.03271},
  year           = {2019},
  doi            = {10.48550/arXiv.1902.03271},
  url            = {https://arxiv.org/abs/1902.03271},
}

@InProceedings{lu2020deep,
  author         = {Lu, Mingyu and Shahn, Zachary and Sow, Daby and Doshi{-}Velez, Finale and Lehman, Li{-}Wei H.},
  title          = {Is Deep Reinforcement Learning Ready for Practical Applications in Healthcare? {A} Sensitivity Analysis of {Duel-DDQN} for Hemodynamic Management in Sepsis Patients},
  booktitle      = {{AMIA} American Medical Informatics Association Annual Symposium},
  year           = {2020},
  month          = nov,
  volume         = {2020},
  pages          = {773--782},
  doi            = {10.48550/arXiv.2005.04301},
}

@Article{wu2023value,
  author         = {Wu, XiaoDan and Li, RuiChang and He, Zhen and Yu, TianZhi and Cheng, ChangQing},
  title          = {A value-based deep reinforcement learning model with human expertise in optimal treatment of sepsis},
  journal        = {npj Digital Medicine},
  year           = {2023},
  month          = feb,
  volume         = {6},
  number         = {1},
  pages          = {15},
  doi            = {10.1038/s41746-023-00755-5},
}

@InProceedings{sun2025exploring,
  author         = {Sun, Yingchuan and Tang, Shengpu},
  title          = {Exploring Time-Step Size in Reinforcement Learning for Sepsis Treatment},
  booktitle      = {RLC 2025 Workshop on Practical Insights into Reinforcement Learning for Real Systems},
  year           = {2025},
  doi            = {10.48550/arXiv.2511.20913},
}

@InProceedings{tallec2019deep,
  author         = {Tallec, Corentin and Blier, L{\'e}onard and Ollivier, Yann},
  title          = {Making deep {Q}-learning methods robust to time discretization},
  booktitle      = "Proceedings of the 36th " # icml,
  series         = {Proceedings of Machine Learning Research},
  year           = {2019},
  month          = jun,
  volume         = {97},
  pages          = {6096--6104},
  doi            = {10.48550/arXiv.1901.09732},
}

@InProceedings{kim2014convolutional,
  author         = {Kim, Yoon},
  title          = {Convolutional neural networks for sentence classification},
  booktitle      = {Proceedings of the 2014 conference on empirical methods in natural language processing ({EMNLP})},
  year           = {2014},
  month          = oct,
  pages          = {1746--1751},
  doi            = {10.3115/v1/D14-1181}
}

@InProceedings{kumar2020conservative,
  author         = {Kumar, Aviral and Zhou, Aurick and Tucker, George and Levine, Sergey},
  title          = {Conservative {Q}-learning for offline reinforcement learning},
  booktitle      = neurips,
  year           = {2020},
  month          = dec,
  volume         = {33},
  pages          = {1179--1191},
  url            = {https://dl.acm.org/doi/10.5555/3495724.3495824},
  urldate        = {2026-02-19},
}

@InProceedings{zhang2021identifying,
  author         = {Zhang,  Kristine and Wang,  Yuanheng and Du,  Jianzhun and Chu,  Brian and Celi,  Leo Anthony and Kindle,  Ryan and Doshi{-}Velez,  Finale},
  title          = {Identifying Decision Points for Safe and Interpretable Reinforcement Learning in Hypotension Treatment},
  booktitle      = {Proceedings of the Machine Learning for Health NeurIPS Workshop},
  series         = {Proceedings of Machine Learning Research},
  year           = {2021},
  month          = dec,
  volume         = {1},
  pages          = {1--9},
  url            = {https://arxiv.org/abs/2101.03309},
  urldate        = {2026-04-08},
}

@InProceedings{lipton2016modeling,
  author         = {Lipton, Zachary C and Kale, David C and Wetzel, Randall and others},
  title          = {Modeling missing data in clinical time series with {RNN}s},
  booktitle      = {Proceedings of the 1st Machine Learning in Health Care},
  series         = {{JMLR} Workshop and Conference Proceedings},
  year           = {2016},
  month          = aug,
  volume         = {56},
  pages          = {253--270},
  doi            = {10.48550/arXiv.1606.04130},
}

@Article{desman2025distributional,
  author         = {Desman,  Jacob M. and Hong,  Zhang-Wei and Sabounchi,  Moein and Sawant,  Ashwin S. and Gill,  Jaskirat and Costa,  Ana C. and Kumar,  Gagan and Sharma,  Rajeev and Gupta,  Arpeta and McCarthy,  Paul and Nandwani,  Veena and Powell,  Doug and Carideo,  Alexandra and Goodwin,  Donnie and Ahmed,  Sanam and Gidwani,  Umesh and Levin,  Matthew A. and Varghese,  Robin and Filsoufi,  Farzan and Freeman,  Robert and Shetreat-Klein,  Avniel and Charney,  Alexander W. and Hofer,  Ira and Chan,  Lili and Reich,  David and Kovatch,  Patricia and Kohli-Seth,  Roopa and Kraft,  Monica and Agrawal,  Pulkit and Kellum,  John A. and Nadkarni,  Girish N. and Sakhuja,  Ankit},
  title          = {A distributional reinforcement learning model for optimal glucose control after cardiac surgery},
  journal        = {npj Digital Medicine},
  year           = {2025},
  month          = may,
  volume         = {8},
  number         = {1},
  pages          = {313},
  doi            = {10.1038/s41746-025-01709-9},
}

@InProceedings{jiang2016doubly,
  author         = {Jiang,  Nan and Li,  Lihong},
  title          = {Doubly Robust Off-policy Value Evaluation for Reinforcement Learning},
  booktitle      = "Proceedings of the 33rd " # icml,
  series         = {Proceedings of Machine Learning Research},
  year           = {2016},
  month          = jul,
  volume         = {48},
  pages          = {652--661},
  url            = {https://proceedings.mlr.press/v48/jiang16.html},
  urldate        = {2026-02-24},
}

@InProceedings{le2019batch,
  author         = {Le, Hoang and Voloshin, Cameron and Yue, Yisong},
  title          = {Batch policy learning under constraints},
  booktitle      = "Proceedings of the 36th " # icml,
  series         = {Proceedings of Machine Learning Research},
  year           = {2019},
  month          = jun,
  volume         = {97},
  pages          = {3703--3712},
  url            = {https://proceedings.mlr.press/v97/le19a.html},
  urldate        = {2026-02-19},
}

@Article{kuo2022health,
  author         = {Kuo, Nicholas I-Hsien and Polizzotto, Mark N and Finfer, Simon and Garcia, Federico and S{\"o}nnerborg, Anders and Zazzi, Maurizio and B{\"o}hm, Michael and Kaiser, Rolf and Jorm, Louisa and Barbieri, Sebastiano},
  title          = {The Health Gym: synthetic health-related datasets for the development of reinforcement learning algorithms},
  journal        = {Scientific data},
  year           = {2022},
  month          = nov,
  volume         = {9},
  number         = {1},
  pages          = {693},
  doi            = {10.1038/s41597-022-01784-7},
}

@Article{plummer2014dysglycaemia,
  author         = {Plummer, Mark P. and Bellomo, Rinaldo and Cousins, Caroline E. and Annink, Christopher E. and Sundararajan, Krishnaswamy and Reddi, Benjamin A. J. and Raj, John P. and Chapman, Marianne J. and Horowitz, Michael and Deane, Adam M.},
  title          = {Dysglycaemia in the critically ill and the interaction of chronic and acute glycaemia with mortality},
  journal        = {Intensive care medicine},
  year           = {2014},
  month          = jul,
  volume         = {40},
  number         = {7},
  pages          = {973--980},
  doi            = {10.1007/s00134-014-3287-7},
}

@Article{adie2023association,
  author         = {Adie, Sarah K. and Ketcham, Scott W. and Marshall, Vincent D. and Farina, Nicholas and Sukul, Devraj},
  title          = {The association of glucose control on in-hospital mortality in the cardiac intensive care unit},
  journal        = {Journal of Diabetes and its Complications},
  year           = {2023},
  month          = apr,
  volume         = {37},
  number         = {4},
  pages          = {108453},
  doi            = {10.1016/j.jdiacomp.2023.108453},
}

@Article{desgrouas2023insulin,
  author         = {Desgrouas, Maxime and Demiselle, Julien and Stiel, Laure and Brunot, Vincent and Marnai, Rémy and Sarfati, Sacha and Fiancette, Maud and Lambiotte, Fabien and Thille, Arnaud W. and Leloup, Maxime and Clerc, Sébastien and Beuret, Pascal and Bourion, Anne-Astrid and Daum, Johan and Malhomme, Rémi and Ravan, Ramin and Sauneuf, Bertrand and Rigaud, Jean-Philippe and Dequin, Pierre-Fran\c{c}ois and Boulain, Thierry},
  title          = {Insulin therapy and blood glucose management in critically ill patients: a 1-day cross-sectional observational study in 69 {French} intensive care units},
  journal        = {Annals of Intensive Care},
  year           = {2023},
  month          = jun,
  volume         = {13},
  number         = {1},
  pages          = {53},
  doi            = {10.1186/s13613-023-01142-9},
}

@Article{robles2021data,
  author         = {Robles Ar{\'e}valo, Aldo and Maley, Jason H and Baker, Lawrence and da Silva Vieira, Susana M and da Costa Sousa, Jo{\~a}o M and Finkelstein, Stan and Mateo-Collado, Roselyn and Raffa, Jesse D and Celi, Leo Anthony and Demichele Iii, Francis},
  title          = {Data-driven curation process for describing the blood glucose management in the intensive care unit},
  journal        = {Scientific data},
  year           = {2021},
  month          = mar,
  volume         = {8},
  number         = {1},
  pages          = {2052--4463},
  doi            = {10.1038/s41597-021-00864-4},
}

@InProceedings{kostrikov2022offline,
  author         = {Kostrikov, Ilya and Nair, Ashvin and Levine, Sergey},
  title          = {Offline reinforcement learning with implicit {Q}-learning},
  booktitle      = iclr,
  year           = {2022},
  month          = apr,
  doi            = {10.48550/arXiv.2110.06169},
}

@Article{zhu2023offline,
  author         = {Zhu, Taiyu and Li, Kezhi and Georgiou, Pantelis},
  title          = {Offline Deep Reinforcement Learning and Off-Policy Evaluation for Personalized Basal Insulin Control in Type 1 Diabetes},
  journal        = {{IEEE} Journal of Biomedical and Health Informatics},
  year           = {2023},
  month          = oct,
  volume         = {27},
  number         = {10},
  pages          = {5087--5098},
  doi            = {10.1109/JBHI.2023.3303367},
}

@Article{nice2009intensive,
  author         = {{NICE-SUGAR Study Investigators}},
  title          = {Intensive versus conventional glucose control in critically ill patients},
  journal        = {New England Journal of Medicine},
  year           = {2009},
  month          = mar,
  volume         = {360},
  number         = {13},
  pages          = {1283--1297},
  doi            = {10.1056/NEJMoa0810625},
}

@Article{gunst2023tight,
  author         = {Gunst,  Jan and Debaveye,  Yves and G\"{u}iza,  Fabian and Dubois,  Jasperina and {De Bruyn},  Astrid and Dauwe,  Dieter and {De Troy},  Erwin and Casaer,  Michael P. and {De Vlieger},  Greet and Haghedooren,  Renata and Jacobs,  Bart and Meyfroidt,  Geert and Ingels,  Catherine and Muller,  Jan and Vlasselaers,  Dirk and Desmet,  Lars and Mebis,  Liese and Wouters,  Pieter J. and Stessel,  Bj\"{o}rn and Geebelen,  Laurien and Vandenbrande,  Jeroen and Brands,  Michiel and Gruyters,  Ine and Geerts,  Ester and {De Pauw},  Ilse and Vermassen,  Joris and Peperstraete,  Harlinde and Hoste,  Eric and {De Waele},  Jan J. and Herck,  Ingrid and Depuydt,  Pieter and Wilmer,  Alexander and Hermans,  Greet and Benoit,  Dominique D. and {Van den Berghe},  Greet},
  title          = {Tight blood-glucose control without early parenteral nutrition in the ICU},
  journal        = {New England Journal of Medicine},
  year           = {2023},
  month          = sep,
  volume         = {389},
  number         = {13},
  pages          = {1180--1190},
  doi            = {10.1056/NEJMoa2304855},
}

@Article{tipirneni2022self,
  author         = {Tipirneni, Sindhu and Reddy, Chandan K.},
  title          = {Self-supervised {Transformer} for sparse and irregularly sampled multivariate clinical time-series},
  journal        = {ACM Transactions on Knowledge Discovery from Data},
  year           = {2022},
  month          = jul,
  volume         = {16},
  number         = {6},
  pages          = {1--17},
  doi            = {10.1145/3516367},
}

@Article{hochreiter1997long,
  author         = {Hochreiter, Sepp and Schmidhuber, J{\"u}rgen},
  title          = {{Long Short-Term Memory}},
  journal        = {Neural Computation},
  year           = {1997},
  month          = nov,
  volume         = {9},
  number         = {8},
  pages          = {1735--1780},
  doi            = {10.1162/neco.1997.9.8.1735},
}

@Article{magni2007model,
  author         = {Magni,  Lalo and Raimondo,  Davide M. and Bossi,  Luca and Dalla Man,  Chiara and De Nicolao,  Giuseppe and Kovatchev,  Boris and Cobelli,  Claudio},
  title          = {Model Predictive Control of Type 1 Diabetes: An in Silico Trial},
  journal        = {Journal of Diabetes Science and Technology},
  year           = {2007},
  month          = nov,
  volume         = {1},
  number         = {6},
  pages          = {804--812},
  doi            = {10.1177/193229680700100603},
}

@Article{bellman1957markovian,
  author         = {Bellman, Richard},
  title          = {A {Markovian} decision process},
  journal        = {Indiana University Mathematics Journal},
  year           = {1957},
  month          = apr,
  volume         = {6},
  number         = {4},
  pages          = {679--684},
  doi            = {10.1512/IUMJ.1957.6.56038},
}

@InProceedings{arnrich2024medical,
  title          = {Medical event data standard ({MEDS}): Facilitating machine learning for health},
  author         = {Arnrich, Bert and Choi, Edward and Fries, Jason Alan and McDermott, Matthew BA and Oh, Jungwoo and Pollard, Tom and Shah, Nigam and Steinberg, Ethan and Wornow, Michael and {van de Water}, Robin},
  booktitle      = iclr # " Workshop on Learning from Time Series For Health",
  pages          = {3--8},
  year           = {2024},
  month          = mar,
  url            = {https://openreview.net/forum?id=IsHy2ebjIG},
  urldate        = {2026-02-19},
}

@InProceedings{raghu2017continuous,
  author         = {Raghu, Aniruddh and Komorowski, Matthieu and Celi, Leo Anthony and Szolovits, Peter and Ghassemi, Marzyeh},
  title          = {Continuous state-space models for optimal sepsis treatment: a deep reinforcement learning approach},
  booktitle      = {Machine learning for healthcare conference},
  year           = {2017},
  volume         = {68},
  month          = aug,
  pages          = {147--163},
  url            = {https://proceedings.mlr.press/v68/raghu17a.html},
}

@Article{johnson2023mimic,
  author         = {Johnson, Alistair E. W. and Bulgarelli, Lucas and Shen, Lu and Gayles, Alvin and Shammout, Ayad and Horng, Steven and Pollard, Tom J. and Hao, Sicheng and Moody, Benjamin and Gow, Brian and Lehman, Li-wei H. and Celi, Leo A. and Mark, Roger G.},
  title          = {{MIMIC-IV}, a freely accessible electronic health record dataset},
  journal        = {Scientific Data},
  year           = {2023},
  month          = jan,
  volume         = {10},
  number         = {1},
  pages          = {1},
  doi            = {10.1038/s41597-022-01899-x},
}

@Misc{johnson2024mimic,
  author       = {Johnson, Alistair and Bulgarelli, Lucas and Pollard, Tom and Gow, Brian and Moody, Benjamin and Horng, Steven and Celi, Leo Anthony and Mark, Roger},
  title        = {{MIMIC-IV} (version 3.1)},
  year         = {2024},
  howpublished = {PhysioNet},
  doi          = {10.13026/kpb9-mt58},
  note         = {RRID:SCR\_007345. Accessed: 2026-02-19},
}

@InProceedings{huang2025mimic,
  author         = {Huang, Yong and Yang, Zhongqi and Rahmani, Amir},
  title          = {MIMIC-Sepsis: A Curated Benchmark for Modeling and Learning from Sepsis Trajectories in the ICU},
  booktitle      = {2025 IEEE EMBS International Conference on Biomedical and Health Informatics (BHI)},
  year           = {2025},
  month          = oct,
  pages          = {1--7},
  doi            = {10.1109/BHI67747.2025.11269536},
}

@InProceedings{voloshin2021empirical,
  author         = {Voloshin, Cameron and Le, Hoang Minh and Jiang, Nan and Yue, Yisong},
  title          = {Empirical Study of Off-Policy Policy Evaluation for Reinforcement Learning},
  booktitle      = {Proceedings of the Neural Information Processing Systems Track on Datasets and Benchmarks},
  year           = {2021},
  month          = dec,
  volume         = {1},
  url            = {https://datasets-benchmarks-proceedings.neurips.cc/paper/2021/file/a5e00132373a7031000fd987a3c9f87b-Paper-round1.pdf},
  urldate        = {2026-02-24},
}

@InProceedings{tang2021model,
  author         = {Tang, Shengpu and Wiens, Jenna},
  title          = {Model selection for offline reinforcement learning: Practical considerations for healthcare settings},
  booktitle      = {Proceedings of the 6th Machine Learning for Healthcare Conference},
  series         = {Proceedings of Machine Learning Research},
  year           = {2021},
  month          = aug,
  volume         = {149},
  pages          = {2--35},
  url            = {https://proceedings.mlr.press/v149/tang21a.html},
  urldate        = {2026-02-19},
  note           = {PMID: 35702420},
}

@Article{gottesman2018evaluating,
  author         = {Gottesman,  Omer and Johansson,  Fredrik and Meier,  Joshua and Dent,  Jack and Lee,  Donghun and Srinivasan,  Srivatsan and Zhang,  Linying and Ding,  Yi and Wihl,  David and Peng,  Xuefeng and Yao,  Jiayu and Lage,  Isaac and Mosch,  Christopher and Lehman,  Li-wei H. and Komorowski,  Matthieu and Komorowski,  Matthieu and Faisal,  Aldo and Celi,  Leo Anthony and Sontag,  David and Doshi-Velez,  Finale},
  title          = {Evaluating reinforcement learning algorithms in observational health settings},
  year           = {2018},
  journal        = {arXiv:1805.12298},
  doi            = {10.48550/arXiv.1805.12298},
}

@InProceedings{marling2020ohiot1dm,
  author         = {Marling, Cindy and Bunescu, Razvan},
  title          = {The {OhioT1DM} dataset for blood glucose level prediction: Update 2020},
  booktitle      = {CEUR Workshop Proceedings},
  year           = {2020},
  month          = sep,
  volume         = {2675},
  pages          = {71--74},
  note           = {PMID: 33584164}
}

@Article{levine2020offline,
  author         = {Levine, Sergey and Kumar, Aviral and Tucker, George and Fu, Justin},
  title          = {Offline reinforcement learning: Tutorial, review, and perspectives on open problems},
  year           = {2020},
  journal        = {arXiv:2005.01643},
  doi            = {10.48550/arXiv.2005.01643},
}

@InProceedings{frank2001simple,
  author         = {Frank, Eibe and Hall, Mark},
  title          = {A simple approach to ordinal classification},
  booktitle      = {European conference on machine learning},
  year           = {2001},
  pages          = {145--156},
  doi            = {10.1007/3-540-44795-4\_13},
}

@InProceedings{hausknecht2015deep,
  author         = {Hausknecht, Matthew J. and Stone, Peter},
  title          = {Deep Recurrent {Q}-Learning for Partially Observable {MDP}s},
  booktitle      = {{AAAI} Fall Symposia},
  year           = {2015},
  month          = nov,
  pages          = {29--37},
  url            = {https://cdn.aaai.org/ocs/11673/11673-51288-1-PB.pdf},
  urldate        = {2026-02-19},
}

@InProceedings{ni2022recurrent,
  author         = {Ni, Tianwei and Eysenbach, Benjamin and Salakhutdinov, Ruslan},
  title          = {Recurrent Model-Free {RL} Can Be a Strong Baseline for Many {POMDP}s},
  booktitle      = "Proceedings of the 39th " # icml,
  series         = {Proceedings of Machine Learning Research},
  year           = {2022},
  month          = jul,
  volume         = {162},
  pages          = {16691--16723},
  url            = {https://proceedings.mlr.press/v162/ni22a.html},
  urldate        = {2026-02-19},
}

@InProceedings{igl2018deep,
  author         = {Igl, Maximilian and Zintgraf, Luisa and Le, Tuan Anh and Wood, Frank and Whiteson, Shimon},
  title          = {Deep variational reinforcement learning for {POMDP}s},
  booktitle      = "Proceedings of the 35th " # icml,
  series         = {Proceedings of Machine Learning Research},
  year           = {2018},
  month          = jul,
  volume         = {80},
  pages          = {2117--2126},
  url            = {https://proceedings.mlr.press/v80/igl18a.html},
  urldate        = {2026-02-19},
}

@Article{white1976procedures,
  author         = {White, Chelsea C},
  title          = {Procedures for the solution of a finite-horizon, partially observed, semi-Markov optimization problem},
  journal        = {Operations Research},
  year           = {1976},
  month          = apr,
  volume         = {24},
  number         = {2},
  pages          = {348--358},
  doi            = {10.1287/opre.24.2.348},
}

@phdthesis{yu2006approximate,
  author         = {Yu, Huizhen},
  title          = {Approximate solution methods for partially observable Markov and semi-Markov decision processes},
  school         = {Massachusetts Institute of Technology},
  year           = {2006},
}

@Article{zhang2016continuous,
  author         = {Zhang, Mimi and Revie, Matthew},
  title          = {Continuous-observation partially observable semi-Markov decision processes for machine maintenance},
  journal        = {IEEE Transactions on Reliability},
  year           = {2016},
  month          = mar,
  volume         = {66},
  number         = {1},
  pages          = {202--218},
  doi            = {10.1109/TR.2016.2626477},
}

@Article{bohe2021individualised,
  author         = {Boh{\'e}, Julien and Abidi, Hassane and Brunot, Vincent and Klich, Amna and Klouche, Kada and Sedillot, Nicholas and Tchenio, Xavier and Quenot, Jean-Pierre and Roudaut, Jean-Baptiste and Mottard, Nicolas and Thiollière, Fabrice and Dellamonica, Jean and Wallet, Florent and Souweine, Bertrand and Lautrette, Alexandre and Preiser, Jean-Charles and Timsit, Jean-François and Vacheron, Charles-Hervé and Hssain, Ali Ait and Maucort-Boulch, Delphine},
  title          = {Individualised versus conventional glucose control in critically-ill patients: the {CONTROLING} study -- a randomized clinical trial},
  journal        = {Intensive Care Medicine},
  year           = {2021},
  month          = sep,
  volume         = {47},
  number         = {11},
  pages          = {1271--1283},
  note           = {PMID: 34590159},
}

@Article{plummer2022time,
  author         = {Plummer, Mark P. and Hermanides, Jeroen and Deane, Adam M.},
  title          = {Is it time to personalise glucose targets during critical illness?},
  journal        = {Current Opinion in Clinical Nutrition \& Metabolic Care},
  year           = {2022},
  month          = sep,
  volume         = {25},
  number         = {5},
  pages          = {364--369},
  doi            = {10.1097/MCO.0000000000000846},
}

@Article{goldberger2000physiobank,
  author         = {Goldberger, Ary L and Amaral, Luis AN and Glass, Leon and Hausdorff, Jeffrey M and Ivanov, Plamen Ch and Mark, Roger G and Mietus, Joseph E and Moody, George B and Peng, Chung-Kang and Stanley, H Eugene},
  title          = {{PhysioBank}, {PhysioToolkit}, and {PhysioNet}: components of a new research resource for complex physiologic signals},
  journal        = {Circulation},
  year           = {2000},
  month          = jun,
  volume         = {101},
  number         = {23},
  pages          = {e215--e220},
  doi            = {10.1161/01.cir.101.23.e215},
}

@Article{liu2020reinforcement,
  author         = {Liu,  Siqi and See,  Kay Choong and Ngiam,  Kee Yuan and Celi,  Leo Anthony and Sun,  Xingzhi and Feng,  Mengling},
  title          = {Reinforcement Learning for Clinical Decision Support in Critical Care: Comprehensive Review},
  journal        = {Journal of Medical Internet Research},
  year           = {2020},
  month          = jul,
  volume         = {22},
  number         = {7},
  pages          = {e18477},
  doi            = {10.2196/18477},
}

@Article{jayaraman2024primer,
  author         = {Jayaraman, Pushkala and Desman, Jacob and Sabounchi, Moein and Nadkarni, Girish N and Sakhuja, Ankit},
  title          = {A primer on reinforcement learning in medicine for clinicians},
  journal        = {npj Digital Medicine},
  year           = {2024},
  month          = nov,
  volume         = {7},
  number         = {1},
  pages          = {337},
  doi            = {10.1038/s41746-024-01316-0},
}

@Article{wang2023optimized,
  author         = {Wang, Guangyu and Liu, Xiaohong and Ying, Zhen and Yang, Guoxing and Chen, Zhiwei and Liu, Zhiwen and Zhang, Min and Yan, Hongmei and Lu, Yuxing and Gao, Yuanxu and Xue, Kanmin and Li, Xiaoying and Chen, Ying},
  title          = {Optimized Glycemic Control of Type 2 Diabetes with Reinforcement Learning: A Proof-of-Concept Trial},
  journal        = {Nature Medicine},
  year           = {2023},
  month          = oct,
  volume         = {29},
  number         = {10},
  pages          = {2633--2642},
  doi            = {10.1038/s41591-023-02552-9},
}

@InProceedings{gao2023offline,
  author         = {Gao, Qitong and Schmidt, Stephen L. and Chowdhury, Afsana and Feng, Guangyu and Peters, Jennifer J. and Genty, Katherine and Grill, Warren M. and Turner, Dennis A. and Pajic, Miroslav},
  title          = {Offline Learning of Closed-Loop Deep Brain Stimulation Controllers for {P}arkinson Disease Treatment},
  booktitle      = {International Conference on Cyber-Physical Systems},
  year           = {2023},
  volume         = {14},
  month          = may,
  pages          = {44--55},
  doi            = {10.1145/3576841.3585925},
}

@Article{fan2026reinforcement,
  author         = {Fan, Fan and Huang, Hao and Yan, Jingwen and Xu, Chao-Yue and Wu, Xiuhua and Zhou, Chunmei and Wen, Dandan and Huang, Hai and Li, Ho Cheung and Qiu, Yihong},
  title          = {Reinforcement learning-based digital therapeutic intervention for postprostatectomy Incontinence: Development and Pilot Feasibility Study},
  journal        = {{JMIR} Cancer},
  year           = {2026},
  month          = feb,
  volume         = {12},
  pages          = {e83375},
  doi            = {10.2196/83375},
}

@InProceedings{tang2020clinician,
  author         = {Tang, Shengpu and Modi, Aditya and Sjoding, Michael W. and Wiens, Jenna},
  title          = {Clinician-in-the-Loop Decision Making: Reinforcement Learning with Near-Optimal Set-Valued Policies},
  booktitle      = "Proceedings of the 37th " # icml,
  series         = {Proceedings of Machine Learning Research},
  year           = {2020},
  month          = jul,
  volume         = {119},
  pages          = {9387--9396},
  url            = {https://dl.acm.org/doi/10.5555/3524938.3525808},
  urldate        = {2026-02-19},
}

@Article{roggeveen2021transatlantic,
  author         = {Roggeveen, Luca and {el Hassouni}, Ali and Ahrendt, Jonas and Guo, Tingjie and Fleuren, Lucas and Thoral, Patrick and Girbes, Armand R. J. and Hoogendoorn, Mark and Elbers, Paul W. G.},
  title          = {Transatlantic transferability of a new reinforcement learning model for optimizing haemodynamic treatment for critically ill patients with sepsis},
  journal        = {Artificial Intelligence in Medicine},
  year           = {2021},
  month          = feb,
  volume         = {112},
  pages          = {102003},
  doi            = {10.1016/j.artmed.2020.102003},
}

@InProceedings{fatemi2021medical,
  author         = {Fatemi, Mehdi and Killian, Taylor W. and Subramanian, Jayakumar and Ghassemi, Marzyeh},
  title          = {Medical Dead-ends and Learning to Identify High-Risk States and Treatments},
  booktitle      = neurips,
  year           = {2021},
  month          = dec,
  volume         = {34},
  pages          = {4856--4870},
  url            = {https://dl.acm.org/doi/10.5555/3540261.3540632},
  urldate        = {2026-02-19},
}

@Article{tu2025offline,
  author         = {Tu, Rui and Luo, Zhipeng and Pan, Chuanliang and Wang, Zhong and Su, Jie and Zhang, Yu and Wang, Yifan},
  title          = {Offline Safe Reinforcement Learning for Sepsis Treatment: Tackling Variable-Length Episodes with Sparse Rewards},
  journal        = {Human-Centric Intelligent Systems},
  year           = {2025},
  month          = feb,
  volume         = {5},
  number         = {1},
  pages          = {63--76},
  doi            = {10.1007/s44230-025-00093-7},
}

@Book{sutton2018reinforcement,
  author         = {Sutton, Richard S and Barto, Andrew G},
  title          = {{Reinforcement Learning: An Introduction}},
  publisher      = {Massachusetts Institute of Technology ({MIT}) Press},
  year           = {2018},
}

@Article{frost2026hidden,
  author         = {Frost, Thomas and Vaidya, Hrisheekesh and Harris, Steve},
  title          = {The hidden risks of temporal resampling in clinical reinforcement learning},
  year           = {2026},
  journal        = {arXiv:2602.06603},
  doi            = {10.48550/arXiv.2602.06603},
}

@Article{mcinnes2018umap,
  author         = {McInnes, Leland and Healy, John and Saul, Nathaniel and Großberger, Lukas},
  title          = {UMAP: Uniform Manifold Approximation and Projection},
  journal        = {Journal of Open Source Software},
  year           = {2018},
  month          = sep,
  volume         = {3},
  pages          = {861},
  doi            = {10.21105/joss.00861},
}

@Article{wilson2007intensive,
  author         = {Wilson, Mark and Weinreb, Jane and {Soo Hoo}, Guy W.},
  title          = {Intensive insulin therapy in critical care: a review of 12 protocols},
  journal        = {Diabetes Care},
  year           = {2007},
  month          = apr,
  volume         = {30},
  number         = {4},
  pages          = {1005--1011},
  doi            = {10.2337/dc06-1964},
}


\newpage

\appendix
\setcounter{table}{0}
\renewcommand{\thetable}{A\arabic{table}}
\setcounter{figure}{0}
\renewcommand{\thefigure}{A\arabic{figure}}

\section{Table of Input Features}

\renewcommand{\arraystretch}{1.5}
\begin{longtable}{>{\bfseries\raggedright\arraybackslash}p{0.25\textwidth} >{\raggedright\arraybackslash}p{0.65\textwidth}}
\caption{Summary of the 140 medical event features available in the patient input state. Unless otherwise specified as oral (PO) or nasogastric (NG), drugs should be assumed to be administered intravenously (IV).}\label{tab:feature-summary} \\
\toprule
Category & Features \\
\midrule
\endfirsthead
\multicolumn{2}{c}%
{{\bfseries \tablename\ \thetable{} -- continued from previous page}} \\
\toprule
Category & \textbf{Features} \\
\midrule
\endhead
\midrule
\multicolumn{2}{r}{{Continued on next page}} \\
\endfoot
\bottomrule
\endlastfoot

Demographics & Age, Gender, Patient Weight \\
\midrule
Lab Tests \& Biomarkers & ALP, ALT, AST, Albumin, Amylase, Anion Gap, Base Excess, Bedside Glucose, Bilirubin, Blood Gas SpO2, Blood Gas pCO2, Blood Gas pO2, CRP, Calcium, Chloride, Creatinine, Glucose, HCO3, Haematocrit, Haemoglobin, Ionised Calcium, LDH, Lactate, Lipase, Platelets, Potassium, Prothrombin Time, Sodium, Troponin - T, Urea, WBC, pH \\
\midrule
Anti-infectives & Aciclovir bolus, Ambisome bolus, Amikacin bolus, Ampicillin bolus, Ampicillin-Sulbactam bolus, Azithromycin bolus, Aztreonam bolus, Caspofungin bolus, Cefazolin bolus, Cefepime bolus, Ceftazidime bolus, Ceftriaxone bolus, Ciprofloxacin bolus, Clindamycin bolus, Co-trimoxazole bolus, Colistin bolus, Daptomycin bolus, Doxycycline bolus, Erythromycin bolus, Gentamicin bolus, Levofloxacin bolus, Linezolid bolus, Meropenem bolus, Metronidazole bolus, Micafungin bolus, Nafcillin bolus, Piperacillin bolus, Piperacillin-Tazobactam bolus, Rifampin bolus, Tigecycline bolus, Tobramycin bolus, Vancomycin bolus, Voriconazole bolus \\
\midrule
Cardiovascular \& Vasoactive & Adrenaline rate, Dobutamine rate, Dopamine rate, Milrinone rate, Noradrenaline rate, Vasopressin rate \\
\midrule
Neurological, Sedatives, Analgesics \& Paralytics & Cisatracurium bolus, Cisatracurium rate, Dexmedetomidine rate, Fentanyl bolus, Fentanyl rate, Ketamine bolus, Ketamine rate, Lorazepam bolus, Lorazepam rate, Midazolam bolus, Midazolam rate, Morphine bolus, Morphine rate, Propofol bolus, Propofol rate, Rocuronium bolus, Rocuronium rate, Vecuronium bolus, Vecuronium rate \\
\midrule
Steroids & Dexamethasone (IV) bolus, Dexamethasone (PO/NG) bolus, Methylprednisolone (IV) bolus, Methylprednisolone (PO/NG) bolus, Prednisolone (PO/NG) bolus \\
\midrule
Diuretics & Bumetanide bolus, Bumetanide rate, Furosemide bolus, Furosemide rate, Mannitol bolus \\
\midrule
Other Medications & Alteplase rate, Aminophylline rate, Amiodarone bolus, Amiodarone rate, IVIG bolus, Labetalol bolus, Labetalol rate, Levetiracetam bolus, Levetiracetam rate, Nitroglycerin rate, Nitroprusside rate, Phenytoin bolus, Phenytoin rate, Sodium Bicarbonate 8.4\% rate, Unfractionated Heparin bolus, Unfractionated Heparin rate \\
\midrule
Fluids, Nutrition \& Blood Products & Dextrose 10\% bolus, Dextrose 10\% rate, Dextrose 20\% rate, Dextrose 5\% bolus, Dextrose 5\% rate, Dextrose 50\% bolus, FFP rate, Hypertonic Saline bolus, Hypertonic Saline rate, Insulin (TPN) rate, PRBC rate, Platelet infusion rate, Regular Insulin bolus, Regular Insulin rate, Carbohydates (PO/NG) rate, Carbohydates (IV) rate, Protein (PO/NG) rate, Protein (IV) rate\\
\midrule
Dialysis & Dialysate Rate, Dialysis Blood Flow Rate, Dialysis Fluid Removal Rate \\
\end{longtable}

\section{Implementation details}\label{app:implementation_details}

All machine learning experiments were conducted using PyTorch and contained with a Jupyter notebook. The model architecture code is shown in Appendix~\ref{app:model_code}. In all instances, a hidden dimension of 64 is used. We used the Adam optimiser with learning rate $1e-3$. Behavioural cloning was trained over 100 epochs; all other experiments were trained over 50 epochs. Critics were trained using target models with Polyak averaging at a rate of 0.01. For IQL, $\tau$ was set to 0.8 and $\beta$ was set to 3.0. For CQL, $\alpha$ was set to 1.0.

\section{Per-episode lengths and frequency of hypo-/hyperglycaemic events}

\begin{table}[h]
  \caption{Summary of per-episode clinical and temporal characteristics in the Insulin4RL dataset.}\label{tab:episode-stats}
  \centering
  \begin{tabular}{lrrrr}
    \toprule
    \textbf{Metric} & \textbf{Duration} & \textbf{Total} & \textbf{Hypoglycaemic} & \textbf{Hyperglycaemic} \\
     & & \textbf{Decisions} & \textbf{Events} & \textbf{Events} \\
     & (hours) & ($n$) & ($<4$ mmol/L) & ($\geq 10$ mmol/L) \\
    \midrule
    Median & 28 & 17 & 0 & 1.0 \\
    Mean & 42 & 28 & 0.4 & 5.8 \\
    Minimum & N/A & 1 & 0 & 0 \\
    Maximum & 840 (35 days) & 598 & 17 & 198 \\
    \bottomrule
  \end{tabular}
\end{table}

\section{Model Architecture Implementation Details}\label{app:model_code}

The following Python code describes the \texttt{CNLSTMModel} architecture and the \texttt{GroupedEmbeddings} module used for processing a sequence of irregularly sampled medical event tuples.

\begin{lstlisting}[language=Python, caption=PyTorch Implementation of the CNN-LSTM Model]
from typing import Union, Tuple
import torch
import torch.nn as nn
import torch.nn.functional as F
from torch.nn.utils.parametrizations import spectral_norm

class GroupedEmbeddings(nn.Module):
    """
    Class for doing one-to-many embeddings for multiple different floats 
    i.e., value, time, etc.
    """
    def __init__(self, n_features: int, n_groups: int, hidden_dim: int = 128, out_dim: int = 128):
        super().__init__()
        self.n_groups = n_groups
        self.n_features = n_features
        self.hidden_dim = hidden_dim

        self.feature_embed = nn.Embedding(n_features, hidden_dim)
        self.norm = nn.LayerNorm(hidden_dim)
        self.conv1 = nn.Conv1d(
                in_channels=n_groups,
                out_channels=n_groups * hidden_dim * 2,
                kernel_size=1,
                groups=n_groups
        )
        self.conv2 = nn.Conv1d(
                in_channels=n_groups * hidden_dim,
                out_channels=n_groups * out_dim,
                kernel_size=1,
                groups=n_groups
        )

    def forward(self, features: torch.Tensor, x: torch.Tensor) -> torch.Tensor:
        N, L, _ = x.shape
        f_emb = self.feature_embed(features.long()).transpose(1, 2).view(N, 1, self.hidden_dim, L)

        x = x.transpose(1, 2)
        x = F.glu(self.conv1(x), dim=1)
        x = x.view(N, self.n_groups, self.hidden_dim, L)
        x = x + f_emb

        # Second projection
        x = x.view(N, -1, L)
        x = self.conv2(x)

        x = x.transpose(1, 2).view(N, L, self.n_groups, -1)
        x = x.sum(dim=2)
        out = self.norm(x)
        return out

class CNNLSTMModel(nn.Module):
    """
    PyTorch model using CNN + LSTM
    """
    def __init__(
            self,
            n_features: int,
            hidden_dim: int = 64,
            n_cnn_layers: int = 2,
            n_lstm_layers: int = 1,
            dropout: float = 0.2,
            out_dim: Union[int, Tuple[int]] = 1
    ):
        super().__init__()
        self.n_cnn_layers = n_cnn_layers
        self.n_lstm_layers = n_lstm_layers

        # Embeddings
        self.embedding_net = GroupedEmbeddings(
            n_features=n_features, n_groups=2, hidden_dim=hidden_dim, out_dim=hidden_dim
        )
        self.embedding_dropout = nn.Dropout1d(dropout)

        # CNN Layers
        cnn_blocks = []
        for _ in range(n_cnn_layers):
            cnn_blocks.extend([
                nn.Conv1d(hidden_dim, hidden_dim, kernel_size=3, stride=1, padding=1),
                nn.GroupNorm(num_groups=8, num_channels=hidden_dim),
                nn.ReLU(),
                nn.MaxPool1d(kernel_size=2, stride=2),
                nn.Dropout1d(dropout),
            ])
        self.cnn = nn.Sequential(*cnn_blocks)

        # LSTM Layer
        self.h0 = nn.Parameter(torch.zeros(n_lstm_layers, 1, hidden_dim))
        self.c0 = nn.Parameter(torch.zeros(n_lstm_layers, 1, hidden_dim))
        nn.init.normal_(self.h0, mean=0.0, std=0.01)
        nn.init.normal_(self.c0, mean=0.0, std=0.01)

        self.lstm = nn.LSTM(
            input_size=hidden_dim, hidden_size=hidden_dim, num_layers=1, batch_first=True
        )

        # Dense Decoding Layers
        if not isinstance(out_dim, tuple):
            out_dim = (out_dim,)

        self.dense_decoder = nn.Sequential(
                nn.Dropout(dropout),
                spectral_norm(nn.Linear(hidden_dim, hidden_dim)),
                nn.LayerNorm(hidden_dim),
                nn.ReLU(),
                nn.Dropout(dropout),
            )

        self.dense_heads = nn.ModuleList([])
        for dim in out_dim:
            self.dense_heads.append(nn.Linear(hidden_dim, dim))

    def soft_update(self, target_model: nn.Module, polyak_tau: float = 0.005):
        with torch.no_grad():
            for param, target_param in zip(self.parameters(), target_model.parameters()):
                target_param.data.lerp_(param.data, polyak_tau)

    def get_lengths_after_conv(self, nan_mask: torch.Tensor) -> torch.Tensor:
        real_mask = (~nan_mask).float()
        for _ in range(self.n_cnn_layers):
            real_mask = F.avg_pool1d(real_mask, kernel_size=2, stride=2)

        real_mask = real_mask.squeeze(1)
        lengths = (real_mask == 1).sum(dim=-1).view(-1, 1, 1) - 1
        return lengths.clamp(min=0)

    def forward(self, x: torch.Tensor) -> torch.Tensor:
        N, L, _ = x.shape
        features, float_inputs = x[..., 0], x[..., 1:]

        # Find the NaNs and mask them
        src_nan_mask = float_inputs[:, :, 0].isnan().view(N, 1, L)
        float_inputs = float_inputs.nan_to_num(nan=0.0)
        features = features.nan_to_num(nan=0.0)

        # Perform embedding
        hidden_state = self.embedding_net(features, float_inputs)

        # Prepare for CNN: (N, L, C) -> (N, C, L)
        hidden_state = hidden_state.transpose(1, 2).contiguous()
        hidden_state = self.embedding_dropout(hidden_state)

        # Perform convolutions
        hidden_state = self.cnn(hidden_state)

        # Prepare for LSTM: (N, C, L') -> (N, L', C)
        hidden_state = hidden_state.transpose(1, 2).contiguous()

        # Update post-cnn NaN mask
        lengths = self.get_lengths_after_conv(src_nan_mask).view(N, 1, 1)

        # Apply LSTM
        h0_expanded = self.h0.expand(self.n_lstm_layers, N, -1).contiguous()
        c0_expanded = self.c0.expand(self.n_lstm_layers, N, -1).contiguous()
        lstm_out, _ = self.lstm(hidden_state, (h0_expanded, c0_expanded))

        # Gather the relevant hidden state based on sequence length
        hidden_state = torch.take_along_dim(lstm_out, lengths, dim=1).squeeze(1)

        # Decode
        hidden_state = self.dense_decoder(hidden_state)
        out = [dense_head(hidden_state) for dense_head in self.dense_heads]
        return out[0] if len(out) == 1 else out
\end{lstlisting}

\end{document}